\documentclass[10pt, a4paper]{article}

\usepackage{lrec-coling2024} 

\usepackage{natbib}
\usepackage{multibib}
\makeatletter
\def\@mb@citenamelist{cite,citep,citet,citealp,citealt,citepalias,citetalias}
\makeatother
\newcites{languageresource}{~}

\usepackage{graphicx}
\usepackage{tabularx}
\usepackage{soul}

\usepackage{CJK}
\usepackage{color}
\usepackage{graphicx}
\usepackage{subfigure}

\usepackage{tabularx}
\usepackage{booktabs}
\usepackage{multirow}
\usepackage{makecell}
\usepackage{tablefootnote}

\usepackage{url}

\usepackage{titlesec}
\titleformat{\section}{\normalfont\large\bfseries\center}{\thesection.}{1em}{}
\titleformat{\subsection}{\normalfont\SmallTitleFont\bfseries\raggedright}{\thesubsection.}{1em}{}
\titleformat{\subsubsection}{\normalfont\normalsize\bfseries\raggedright}{\thesubsubsection.}{1em}{}
\renewcommand\thesection{\arabic{section}}
\renewcommand\thesubsection{\thesection.\arabic{subsection}}
\renewcommand\thesubsubsection{\thesubsection.\arabic{subsubsection}}

\usepackage{xcolor}
\usepackage{hyperref}
 \definecolor{darkblue}{rgb}{0, 0, 0.5}
  \hypersetup{colorlinks=true, citecolor=darkblue, linkcolor=darkblue, urlcolor=darkblue}

\usepackage{xstring}

\usepackage{color}

\title{An In-depth Survey of Large Language Model-based Artificial Intelligence Agents}

\name{Pengyu Zhao$^\ast$, Zijian Jin$^\ast$, Ning Cheng} 

\address{Beijing Jiaotong University, New York University, \\
          zj2076@nyu.edu\\
         \{pengyuzhao, ningcheng\}@bjtu.edu.cn\\}

\abstract{
Due to the powerful capabilities demonstrated by large language model (LLM), there has been a recent surge in efforts to integrate them with AI agents to enhance their performance. In this paper, we have explored the core differences and characteristics between LLM-based AI agents and traditional AI agents. Specifically, we first compare the fundamental characteristics of these two types of agents, clarifying the significant advantages of LLM-based agents in handling natural language, knowledge storage, and reasoning capabilities. Subsequently, we conducted an in-depth analysis of the key components of AI agents, including planning, memory, and tool use. Particularly, for the crucial component of memory, this paper introduced an innovative classification scheme, not only departing from traditional classification methods but also providing a fresh perspective on the design of an AI agent's memory system. We firmly believe that in-depth research and understanding of these core components will lay a solid foundation for the future advancement of AI agent technology. At the end of the paper, we provide directional suggestions for further research in this field, with the hope of offering valuable insights to scholars and researchers in the field.
 \\ \newline  }

\begin{document} 

\maketitleabstract
\renewcommand{\thefootnote}{[]}
\renewcommand{\thefootnote}{\fnsymbol{footnote}}
\footnotetext[1]{Equal contribution.}
\renewcommand{\thefootnote}{\arabic{footnote}}
\section{Introduction}
\noindent The notion of intelligent agents can trace its roots back to the research of the mid to late 20th century. Pioneering contributions in this realm encompass Hewitt's Actor model~\cite{hewitt1973universal} and Minsky's innovative conceptualization in the 'Society of Mind'~\cite{Minsky88} which still trigger some new ideas recently eg: "Mindstorms in Natural Language-Based Societies of Mind" ~\cite{zhuge2023mindstorms}.In the 1990s, Russell introduced the framework for intelligent and rational agents ~\cite{russel2010}, which has since become a foundational theory in this field.
The advent of deep neural networks post-2012 marked a significant shift in the AI landscape. Leveraging the power of backpropagation~\cite{rumelhart1986learning} for training deep models, researchers began to explore more sophisticated agent behaviors, transcending beyond traditional rule-based methods. Among the emergent methodologies, Reinforcement Learning (RL) stood out as a paradigm where agents learn optimal behavior through interactions with the environment and receiving feedback in the form of rewards or penalties. In 2013, DeepMind ~\cite{DBLP:journals/corr/MnihKSGAWR13} used RL to play the Atair Game and win humans' performance which indicates that AI Agents are available to outperform human capabilities in specific areas. The incorporation of neural networks into RL, often referred to as Deep Reinforcement Learning (DRL) ~\cite{li2017deep}, allowed for the tackling of previously intractable problems, bridging the gap between high-dimensional input spaces and complex decision-making processes~\cite{arulkumaran2017deep}. 
Despite the promising advancements offered by DRL, certain challenges persist. Chief among these is the issue of generalization. Many reinforcement learning agents, especially those trained in simulated environments, struggle to transfer their learned behavior to new or slightly altered scenarios, often termed as domain adaptation~\cite{arndt2020meta}. Training these agents can also be computationally intensive, often requiring vast amounts of interactions to achieve satisfactory performance. Furthermore, Reinforcement learning training struggles with convergence and the design of reward functions can be challenging, particularly in real-world scenarios, and can be a daunting and often unfeasible task. This hampers the rapid development and deployment of RL-based agents in diverse environments.

In 2020, OpenAI released GPT3~\cite{brown2020language} with 175 billion parameters, making it the largest publicly available language model at the time.  These models, characterized by their immense size and capacity, have shown exceptional prowess in generalization across a myriad of tasks. The ability of LLMs to understand and generate language allows them to act as a foundational model for a wide range of applications ~\cite{huang2022towards}. Their inherent generalization capabilities make them ideal candidates to serve as base models for universal agents. By harnessing the vast knowledge embedded within LLMs, researchers are now exploring hybrid models, integrating the strengths of reinforcement learning with the generalization capacities of LLMs ~\cite{hu2023enabling}. This symbiotic combination promises to pave the way for more robust, adaptable, and efficient intelligent agents in the future.

In order to assist readers in quickly understanding the research history of AI agents and to further inspire research in AI agents, in this paper, we offer a comprehensive and systematic review of AI agents based on the components\footnote{The key components of AI agents were originally defined at https://lilianweng.github.io/posts/2023-06-23-agent/} and applications.

\section{LLM vs. Traditional Agents}
Traditional agents were designed specifically to address certain problems. They primarily relied on predetermined algorithms or rule sets, excelling in tasks they were built for. However, they often struggled with generalization and reasoning when confronted with tasks outside their initial scope.

The introduction of Large Language Models (LLMs) has brought significant changes to AI agent design. These agents, trained on the extensive corpus, are not only proficient in understanding and generating natural language but also display strong generalization abilities. This capability allows them to easily integrate with various tools, enhancing their versatility. On the other hand, the emergent abilities of Large Language Models~\cite{wei2022emergent} shows that LLMs are also good at reasoning which can help them learn from fault behavior.

Taking game exploration as an example, especially in the Minecraft setting, the differences between LLM-based agents like VOYAGER~\cite{wang2023voyager} and traditional RL agents are evident. LLM agents, with their rich pre-trained knowledge, have an advantage in decision-making strategies even without task-specific training. On the other hand, traditional RL agents often need to start from scratch in new environments, relying heavily on interaction to learn. In this scenario, VOYAGER showcases better generalization and data efficiency.

\section{Components of AI Agents}
\subsection{Overview}
The LLM-powered AI agent system relies on LLM to function as its brain, which is supported by several crucial components that deploy various important functions. These functions, including planning, memory, and tool use, have been studied independently and thoughtfully in the past and have a well-established history. In this survey, we will introduce the research history of each individual functional model, mainstream methods, combination methods with the AI agent, and potential directions for the future. We hope that this historical information will serve as an inspiration for the future development of AI agents. It is worth noting that the integration of these three functional models is still a relatively new concept.
\subsection{Planning}

\begin{figure*}[htbp]%
    \begin{center}
    \includegraphics[width=0.9\textwidth]{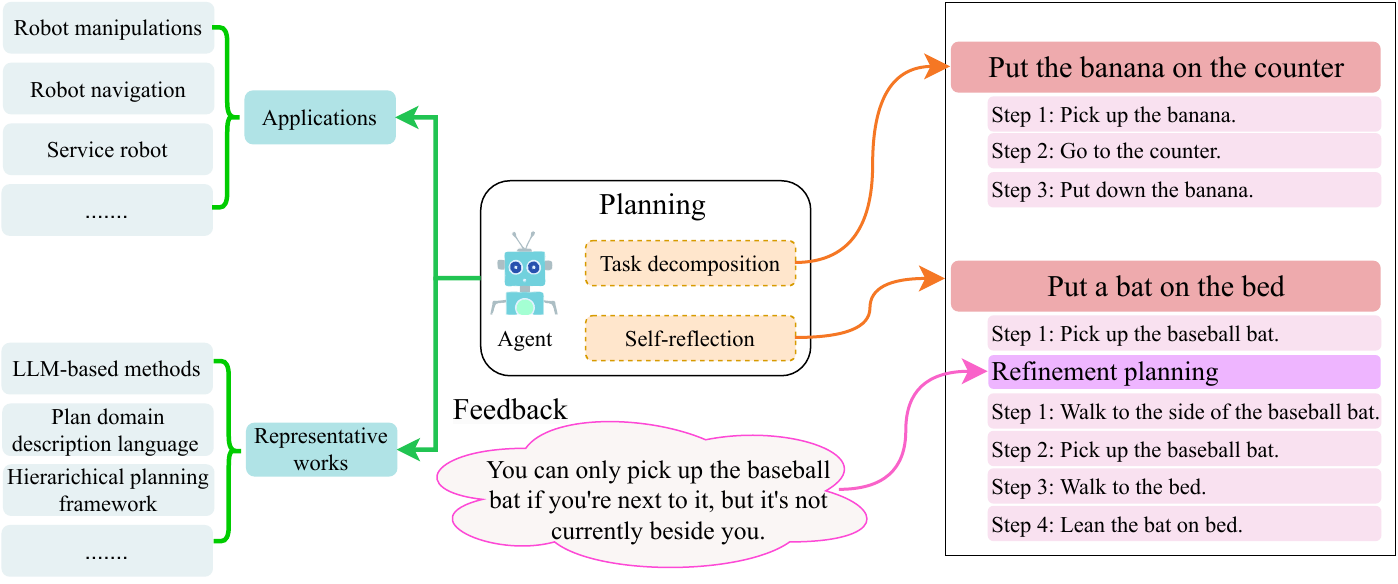}
    \caption{Overview of the planning component of AI agent. \textit{Left} introduces some applications and representative methods of planning. \textit{Right} provides an example illustrating the working mechanism of an AI agent with task decomposition and self-reflection.}
    \label{fig:planningv2}
    \end{center}
\end{figure*}

The goal of planning is to design a series of actions to facilitate state transitions and ultimately achieve the desired task. As shown in the left of Figure \ref{fig:planningv2}, 
this component, functioning as an individual module, has been integrated in various applications, such as robot manipulations~\cite{chen2021optimal}, robot navigation~\cite{lo2018petlon}, and service robots~\cite{li2023adaptive}. And the existing works, such as methods using the planning domain description language (PDDL)~\cite{aeronautiques1998pddl,fox2003pddl2,jiang2019task} and hierarchical planning frameworks~\cite{erol1994htn,suarez2018interleaving,guo2023recent}, have greatly propelled the advancement of planning systems. 
Recently, with significant successes achieved by LLMs in various domains, numerous studies have been exploring the utilization of LLMs to enhance the planning and execution capabilities of AI agents.  
Benefiting from the powerful inference capabilities of LLM, 
LLM-based AI agents can efficiently decompose complex tasks or instructions into a series of sub-tasks or simpler instructions (i.e., planning). For instance, as shown in the top right of Figure \ref{fig:planningv2}, the LLM-based agent decomposes the complex instruction ``Put the banana on the counter'' into a series of simpler instructions which are easier for the agent to accomplish.
Further, taking actions solely based on the initial plan formulated by the agent without considering external environmental feedback may limit the performance of the agent. For example, as shown in the bottom right of Figure \ref{fig:planningv2}, an agent creates a plan for the instruction ``Put the bat on the bed'', and the first step in the initial planning is ``Pick up the baseball bat'', which may fail to execute when there is no 'bat' nearby. However, if the agent can self-reflection based on the feedback, it can refine the first step to "Walk to the side of the baseball bat", and then progressively work towards achieving the goal.
Therefore, during the execution process, reflecting on and analyzing past behaviors and feedback, and subsequently adjusting the plan, are equally pivotal for the successful execution of tasks by AI agents.
Next, we will introduce relevant works that utilize LLM for task decomposition and self-reflection. 

\subsubsection{Task Decomposition}
Task decomposition aims to decompose the complex task or instruction into a series of simpler sub-goals or sub-instructions for performing the task. For example, as shown in the top right of Figure \ref{fig:planningv2}, given a task instruction "Put the banana on the counter", the agent will split it into three steps: 1. Pick up the banana. 2. Go to the counter. 3. Put down the banana. The existing works mainly perform task decomposition by chain or tree of thought~\cite{wei2022chain,kojima2022large,yao2023tree} and PDDL with LLM~\cite{liu2023llm+}. 
Chain of thought can utilize a few examples or simple instructions to progressively guide LLM reasoning, in order to decompose complex tasks into a series of simpler tasks~\cite{wei2022chain,zhang2022automatic,huang2022language,WangXLHLLL23}. Zhang et al.~\cite{zhang2022automatic} proposed a method for automatically generating chain of thought samples. They first clustered the problems and then, for each cluster, selected representative questions to generate chain of thought samples in a zero-shot manner. Huang et al.~\cite{huang2022language} utilized high-level tasks related to the given task and their decomposed planning steps as examples, and combined these examples with input information to construct prompts. Then, they employed LLM to predict the next steps of planning and added the generated steps to the original prompts, continuing the prediction until the entire task was completed. Wang et al.~\cite{WangXLHLLL23} proposed that by guiding LLM to first construct a series of plans and then progressively execute solutions, it can effectively alleviate the issue of intermediate plans disappearing during the reasoning process. 
Unlike linear thinking, the Tree of Thought \cite{long2023large,yao2023tree} generates multiple branches of thoughts at each step to create a tree-like structure. Subsequently, searching on this tree of thought is conducted using methods like breadth-first search or depth-first search. For evaluating each state, reasoning can be facilitated using a "value prompt" or assessment results can be generated through a voting mechanism.
In addition, some research efforts consider combining LLM with PDDL for the purpose of planning target problems~\cite{xie2023translating,liu2023llm+,guan2023leveraging}. For example, Liu et al.~\cite{liu2023llm+} first conveyed the task description in the form of natural language to LLM for translating to PDDL format by in-context learning, then they employed the classical planners to generate plans and converted them into natural language format by LLM again.
\subsubsection{Self-Reflection}
During the process of interacting with the environment, AI agents can enhance their planning ability by reflecting on past actions by receiving feedback. There are many works attempt to combine LLM-based agents with the self-reflection~\cite{yao2022react,huang2022inner,shinn2023reflexion,liu2023chain,sun2023adaplanner,singh2023progprompt,yao2023retroformer,chen2023interact}. For example, Yao et al.~\cite{yao2022react} integrated actions with the chain of thought, leveraging thought to formulate planning that guides the agent's execution of acts. Simultaneously, interactive execution of actions in the environment further enhances the agent's planning ability. Shinn et al.~\cite{shinn2023reflexion}
introduced a framework named Reflexion, in which the approach first generates actions through the Actor module and evaluates them. Then utilizes the self-reflection module to generate feedback and store it in memory. When errors occur, this method can infer the actions that led to the errors and correct them, thereby continuously enhancing the agent's capabilities. Liu et al.~\cite{liu2023chain} first rated the various outputs of the model based on human feedback, then they used prompt templates to construct these ratings into natural language forms and combined them with the outputs for fine-tuning the model, thereby enabling it to learn self-reflection. Singh et al.~\cite{singh2023progprompt} utilize Pythonic program and annotations to generate planning, wherein assertion functions are used to obtain feedback from the environment. When assertions are false, error recovery can be performed. Sun et al.~\cite{sun2023adaplanner} proposed a model named AdaPlanner, which utilizes two refiners to optimize and refine plans. One of the refiners collects information from the environment after executing an action, which is then utilized for subsequent actions. The other one adjusts the existing plan based on feedback obtained from the external environment when the executed action fails to achieve its intended outcome. Similarly, Yao et al~\cite{yao2023retroformer}. first finetuned a small language model as a retrospective model to generate feedback for past failures, and then appended this feedback to the actor prompt as input of the large LLM for preventing the recurrence of similar errors and predicting the next action.

\subsection{Memory}
Memory can help individuals integrate past learned knowledge and experience events with their current state, thereby assisting in making more appropriate decisions.
In general, human memory can be categorized into three primary types: sensory memory, short-term memory, and long-term memory~\cite{camina2017neuroanatomical}. Sensory memory is the collection of information through the senses of touch, hearing, vision, and other senses, and it has an extremely brief lifespan~\cite{wan2020artificial,jung2019bioinspired}. 
Short-term memory refers to the process of handling information within a brief period, and it is typically carried out by working memory~\cite{hunter1957memory,baddeley1983working,baddeley1997human}. 
In contrast, long-term memory refers to memories that can be stored for an extended period, which encompasses episodic memory and semantic memory. Episodic memory refers to the memory capacity for events that individuals have personally experienced, and it is often able to closely associate these events with contextual information~\cite{tulving1972episodic,tulving1983elements}. 
Semantic memory refers to the factual knowledge that individuals know, and this type of memory is unrelated to specific events and personal experiences~\cite{tulving1972episodic}.

\begin{figure*}[htbp]%
    \begin{center}
    \includegraphics[width=0.80\textwidth]{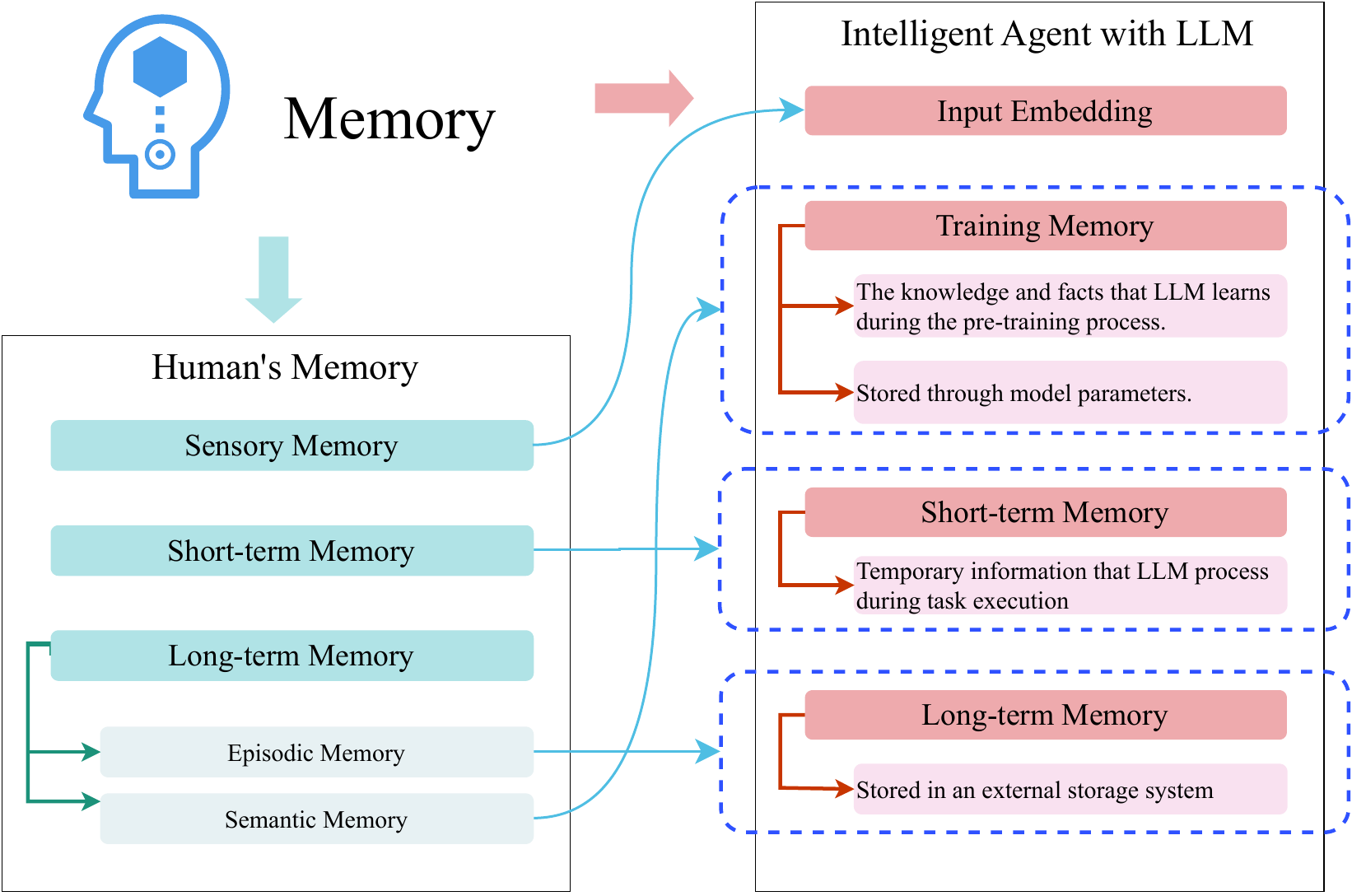}
    \caption{Mapping Structure of Memory: \textit{Left} illustrates memory categories in human memory, while the \textit{right} depicts memory categories in AI agents, which have been redefined based on the characteristics of LLM.}
    \label{fig:memory}
    \end{center}
\end{figure*}

Similarly, memory, as a key component of AI agents, can assist them in learning valuable knowledge from past information, thereby helping the agents perform tasks more effectively. 
To fully utilize the stored information in memory, some research has attempted to integrate AI agents with short-term memory~\cite{kang2023think,peng2023check}, long-term memory~\cite{vere1990basic,kazemifard2014emotion}, and a combination of both~\cite{nuxoll2007extending,kim2023machine,yao2023retroformer,shinn2023reflexion}. In addition, since sensory memory can be regarded as the embedded representation of inputs such as text and images, similar to a sensory buffer, we consider sensory memory not to be part of the memory module of the AI agent.
With the emergence of large language models (LLM), some works devoted to drive the development of AI agents using LLM. Considering the characteristics of LLM, as shown in Figure \ref{fig:memory}, we further redefine the concepts of memory types for AI agents and classify them into training memory, short-term memory, and long-term memory.

Training memory refers to the knowledge and facts that a model learns during the pre-training process, and this information is stored through model parameters. Existing research has shown that models can learn world knowledge~\cite{rogers2021primer}, relational knowledge~\cite{petroni2019language,safavi2021relational}, common sense knowledge~\cite{davison2019commonsense,da2021analyzing,bian2023chatgpt}, semantic knowledge~\cite{tang2023graspgpt}, and syntactic knowledge~\cite{chiang2020pretrained} during the pre-training phase. 
Therefore, by employing LLM for reasoning, the AI agent can implicitly recall this knowledge to enhance the model's performance. 

Short-term memory refers to the temporary information that AI agents process during task execution, such as the example information involved in the in-context learning process and the intermediate results generated during LLM inference. During the inference process, LLM  temporarily stores and processes in-context information or intermediate results, using them to improve the ability of the model. This is similar to human working memory, which temporarily holds and processes information in the short-term to support complex cognitive tasks~\cite{gongworking}. Some works utilize in-context learning to improve the performance of LLM. They first combine some examples with input information to construct a prompt and then send this prompt to LLM to utilize short-term memory~\cite{li-etal-2023-large,logeswaran2022few,omidvar2023empowering}. 
For example,
Li et al.~\cite{li-etal-2023-large} pointed out that when provided with a context that is relevant to the task, it is important to ensure that its working memory is controlled by the context. Otherwise, the model should rely on the world knowledge obtained during the pre-training phase. Logeswaran et al.~\cite{logeswaran2022few} first combined some examples with input instructions as a prompt, and then generated multiple candidate sub-goal plans using LLM. Subsequently, they employed a re-rank model to select the most suitable plan from these candidates.
Some works prompt LLM to output its thinking process and results in the form of chain-of-thought, or to feed the intermediate results from LLM's inference into LLM for further reasoning~\cite{huang2022language,akyurek-etal-2023-rl4f,chen2023introspective,chen2023you,zhang2023investigating,chen2023chatcot}. For example, Zhang et al.~\cite{zhang2023investigating} first guided the model to generate a chain of thought by engaging it in multi-turn dialogues based on the given context. Subsequently, they combined the context with the generated chain of thought to form samples, which are then used to assist the model in reasoning and prediction under new contextual situations. Akyurek et al.~\cite{akyurek-etal-2023-rl4f} proposed a multi-agent collaborative system that includes two LLMs. One LLM is responsible for generating answers based on the input content, while the other LLM generates a textual critique based on the input and output of the first LLM to assist in error correction.

Long-term memory refers to the information stored in an external storage system, and when AI agents use this memory, they can retrieve information relevant to the current context from the external storage. The utilization of long-term memory can be divided into three steps: information storage, information retrieval, and information updating. Information storage aims to store essential information from the interactions between the agent and its environment. For example, Shuster et al.~\cite{shuster2022blenderbot} first generated a summary of the last interaction. If the generated summary is "no persona," it is not stored; otherwise, the summary information is stored in long-term memory. Zhang et al.~\cite{zhang2023large} utilized a tabular format to store memory in the form of key-value pairs. In this format, the observations and states serve as the keys, and the actions and their corresponding Q-values are stored as values. 
Liang et al.~\cite{liang2023unleashing} stored the relevant information from the interactions between the agent and the environment. The information from the last interaction is stored in the flash memory for quick retrieval. The rest of the information is stored in the action memory as long-term memory.
Information retrieval aims to retrieve information relevant to the current context from long-term memory to assist the agent in performing tasks. For example, Lee et al.~\cite{lee2023prompted} first clarified the input information, then they employed dense passage retrievers to select relevant information from long-term memory. Afterward, they combined the selected information with the input information and used methods like chain-of-thought or few-shot learning to choose the most relevant information for task execution. Zhang et al.~\cite{zhang2023large} first computed the similarity between the received information and the keys stored in the long-term memory, and then selected the top k records with the highest similarity to assist the LLM's decision-making. 
Information updating aims to update the stored long-term memory. For example, Zhong et al.~\cite{zhong2023memorybank} designed a forgetting mechanism based on the Ebbinghaus forgetting curve to simulate the updating process of human long-term memory.

\subsection{Tool Use}
Recent works have greatly propelled the development of LLMs, however, LLMs still fail to achieve satisfactory performance in certain scenarios involving up-to-date information, computational reasoning, and others. For example, when a user asks, 'Where is the global premiere of Oppenheimer?', ChatGPT is unable to answer this question because the movie 'Oppenheimer' is the latest information and is not included in the training corpus of the LLM.

To bridge these gaps, many efforts have been dedicated to integrating LLM with external tools to extend its capabilities. Some works aim to integrate LLM with specific tools such as web search~\cite{nakano2021webgpt}, translation~\cite{thoppilan2022lamda},  calculators~\cite{cobbe2021training}, and some plugins of ChatGPT\footnote{https://openai.com/blog/chatgpt-plugins}. 
Some other works consider teaching LLMs to choose suitable tools or combine various tools to accomplish tasks. For example, 
Karpas et al.~\cite{karpas2022mrkl} implemented a system named MRKL, which mainly consists of a language model, an adapter, and multiple experts (e.g., model or tools), where the adapter is utilized to select the appropriate expert to assist the language model in processing input requests. Parisi et al.~\cite{parisi2022talm} designed an iterative self-play algorithm to assist LM in learning how to utilize external APIs by fine-tuning LM. In self-play, they first fine-tuned LM with a few samples and then utilized it to generate the tool input for invoking the tool API to generate results, followed by an LM to infer an answer. If the referred answer is similar to the golden answer, the task input and predicted results (i.e., tool input, tool result, and predicted answer) are appended to the corpus sets for further fine-tuning and iteration in the next round.
Patil et al.~\cite{patil2023gorilla} first constructed a dataset with the format of instruct-API pairs, and then fine-tuned LLM based on the dataset for aiding LLM to employ tools with zero-shot and retriever-aware.   
Similarly, Schick et al.~\cite{schick2023toolformer} fine-tuned the LLM on a dataset containing API calls to help the LLM learn the ability to invoke APIs.
Paranjape et al.~\cite{paranjape2023art} first retrieved the related examples with the input task as a prompt and then employed the LLM to implement inference with chain reasoning. In this process, if the immediate step requires tools, the inference process is paused to execute the tools, and the output of the tools is inserted into the inference process.
Li et al.~\cite{li2023api} proposed the API bank to evaluate the LLM's ability to utilize tools and devised a tool-augmented LLM paradigm to alleviate the limitation of in-context length.
Shen et al.~\cite{shen2023hugginggpt} proposed a method to combine LLM with HuggingFace to enhance the performance of LLM. Specifically, the method first employs LLM to decompose complex tasks into a series of sub-tasks and then sequentially selects suitable models from HuggingFace to perform these sub-tasks.
Lu et al.~\cite{lu2023chameleon} designed a plug-and-play compositional reasoning method, which first plans the schedule of input tasks and then composes multiple tools to execute sub-tasks for achieving the original task. 
Liang et al.~\cite{liang2023taskmatrix} first applied a multi-model foundation model to understand and plan the given instructions for selecting suitable APIs from the API platform, and then utilized an action executor to generate results based on the selected APIs. Besides, they also exploited the feedback of humans to optimize the ability of planning and choose APIs of LLM, and the document of API in API platform.
Different from the above approaches, Cai et al.~\cite{cai2023large} first employed an LLM to generate tool for input task, and then utilized an LLM to perform task based on the generated tool. Specifically, for an incoming task, if the tool required by the task has been generated, the tool will be invoked directly, otherwise, the LLM will first generates tool, and then uses it.

\begin{table*}[!ht]
\scriptsize
\centering
\begin{tabular}{ccp{8cm}}
\toprule
 \textbf{Category } & \textbf{Application} & \makecell[c]{\textbf{Description}}\\ 
         \midrule
 \makecell[c]{Chatbot} & Pi &  Inflection's chatting AI agent known for its emotional companionship and high emotional intelligence \\ 
         \midrule
 Game & Voyager~\cite{wang2023voyager} &  The first LLM-powered embodied lifelong learning agent in Minecraft that continuously explores the world, acquires diverse skills, and makes novel discoveries without human intervention \\ 
        \midrule
Coding  & GPT Engineer&  A AI coding agent that can generate an entire codebase based on a prompt \\ 
       \midrule
 Design & Diagram &  An AI-powered and automatable design platform \\ 
        \midrule
 \multirow{2}{*}{Research} 
    & ChemCrow~\cite{bran2023chemcrow} & An LLM chemistry agent designed
to accomplish tasks across organic synthesis, drug discovery, and materials design \\
    & Agent~\cite{boiko2023emergent} &  An intelligent agent system that combines multiple large language models for autonomous
design, planning, and execution of scientific experiments \\ 
    
        \midrule
 \multirow{5}{*}{Collaboration} 
    & DialOp~\cite{lin2023decision} & AI assistants 
    collaborating with one or more humans via natural language to help them make complex decisions \\
    & MindOS & An engine creating autonomous AI agents for users' professional tasks \\
    & MetaGPT & An multi-agent framework assigning different roles to GPTs to form a collaborative software entity for complex tasks \\
    & Multi-GPT & An experimental multi-agent system where multiple ``expertGPTs" collaborate to perform a task and each has their own short and long-term memory and the ability to communicate with each other. \\
    & Generative Agents~\cite{park2023generative} &  Multiple AI agents for the interactive simulacra of human behavior \\     
        \midrule
 \multirow{6}{*}{General purpose} 
    & Auto-GPT&  An AI agent chaining LLM ``thoughts" together to autonomously achieve whatever goal users set \\ 
    & BabyAGI & An task-driven autonomous agent leveraging GPT-4 language model, Pinecone vector search, and the LangChain framework to perform a wide range of tasks across diverse domains \\
    & SuperAGI & A developer-centric open-source framework to build, manage and run useful Autonomous AI Agents  \\
    & AgentGPT   & A framework allow users to configure and deploy Autonomous AI agents rapidly  \\
\bottomrule
\end{tabular}
\caption{LLM-based AI Agent applications.}
\label{tab:application}
\vspace{-1.5em}
\end{table*}

\section{Application}
  AI Agent is not an emergent concept. As early as 1959, the world's first complete artificial intelligence system, \emph{advice taker}~\cite{mccarthy1959programs}, was proposed. Subsequently, John McCarthy and others began to use the term \emph{Agent} to describe the role that a computing program can play in a scene to achieve certain tasks in artificial intelligence. With reinforcement learning coming into prominence, the field of artificial intelligence has seen a number of notable AI agents based on reinforcement learning and gaming strategies, such as \emph{AlphaGo}~\cite{silver2016mastering}, a Go agent launched by DeepMind in 2014. Similarly, OpenAI launched \emph{OpenAI Five}~\cite{berner2019dota} for playing the game of Dota 2 in 2017 and DeepMind announced \emph{AlphaStar}~\cite{vinyals2019grandmaster} for playing StarCraft II. Recently, the emergence of ChatGPT has made AI agents active once again. The LLM-based Agent also keeps emerging. In this paper, we focus on the latest LLM-based AI Agent applications and talk about the applications of AI Agent from seven aspects: chatbot, game, design, research, coding, collaboration, and general purpose, as shown in Tab. \ref{tab:application}.

\subsection{Chatbot}
Pi\footnote{\url{https://pi.ai/talk}} is a typical LLM-based chatting AI agent released by Inflection. Like ChatGPT\footnote{\url{https://chat.openai.com}} and Claude\footnote{\url{https://www.anthropic.com/index/claude-2}}, users can talk directly with Pi, but Pi not only serves productivity needs such as searching or answering questions but also focuses on emotional companionship. Pi is known for its high emotional intelligence. Users can communicate with Pi as naturally as they would with a close friend.

\subsection{Game}
No other LLM-based gaming intelligence has recently received more attention than Voyager~\cite{wang2023voyager}. Voyager is an AI agent with access to GPT-4~\cite{openai2023gpt4}. Voyager shows remarkable proficiency in playing the game of Minecraft and is able to utilize a learned skill library to solve new tasks from scratch without human intervention, demonstrating strong in-context lifelong learning capabilities.

\subsection{Coding}
Developers have always wanted to have a code generator to help improve programming efficiency. LLM-based agents are naturally used in code generation. A very attractive coding agent is GPT Engineer\footnote{\url{https://github.com/AntonOsika/gpt-engineer}}, which can generate an entire codebase according to a prompt. GPT Engineer even learns the developer's coding style and lets the developer finish the coding project in just a few minutes. What makes GPT Engineer unique is that GPT Engineer asks many detailed questions to allow developers to clarify missing details instead of accepting these requests unconditionally made by developers.

\subsection{Design}
The idea of AI Agent has also been applied to design. Diagram\footnote{\url{https://diagram.com/}} is a representative AI-powered and automatable design platform with many products, including Magician, Genius, Automator, and UI-AI, for designing high-quality charts and graphs. Taking Genius and UI-AI as examples. Genius is equivalent to a design assistant, helping to transform users' ideas into designs. Users only need to provide a product description and Genius can create fully editable UI designs. In addition, Genius can provide design suggestions to help improve productivity. UI-AI contains a series of user interface AI models made for designers that leverage the latest advancements in AI combined with creative prompting or multimodal prompts to generate design assets.

\subsection{Research}
A number of AI agents for autonomous scientific research have emerged. ChemCrow~\cite{bran2023chemcrow} is an LLM chemistry agent designed to accomplish various tasks such as organic synthesis, drug discovery, and materials design. It integrates 17 expert-designed chemistry tools and operates by prompting GPT-4 to provide specific instructions about the task and the format required. Specifically, a set of tools is created by using a variety of chemistry-related packages and software. These tools and user prompts are provided to GPT-4 and GPT-4 determines its behavioral path before arriving at the final answer through an automated, iterative chain-of-thought process. Throughout the process, ChemCrow serves as an assistant to expert chemists while simultaneously lowering the entry barrier for non-experts by offering a simple interface to access accurate chemical knowledge.

Agent~\cite{boiko2023emergent} is an exploration of emerging autonomous scientific research capabilities of large language models. It binds multiple LLMs together for autonomous design, planning, and execution of scientific experiments (eg., the synthesis experiment of ibuprofen and the cross-coupling experiment of Suzuki and Sonogashira reaction). Specifically, autonomous scientific research is accomplished through a series of tools for surfing the Web, reading documents, executing code, etc., and several LLMs for well-timed calls.

\subsection{Collaboration}
Collaboration is one of the most significant applications of AI agents. Many researchers have already started to develop the application by allowing different AI agents to collaborate with each other, such as AI lawyers, AI programmers, and AI finance to form a team to complete complex tasks together. DialOp~\cite{lin2023decision} describes a simple collaborative morphology, in which AI assistants collaborate with one or more humans via natural language to help them make complex decisions. The autonomous AI agents currently created by MindOS\footnote{\url{https://mindos.com/marketplace}} are also used for simple human-agent collaboration to assist users with professional tasks. Compared to DialOp and MindOS, MetaGPT\footnote{\url{https://github.com/geekan/MetaGPT}}and Multi-GPT\footnote{\url{https://github.com/sidhq/Multi-GPT}} allow multiple agents can automatically divide up the work and collaborate with each other to accomplish a task, with MetaGPT focusing more on software industry tasks. 

Additionally, Generative Agents~\cite{park2023generative} are introduced to simulate human behavior. By extending LLMs, complete records of the experiences of the generative agents are stored using natural language, and over time these memories are synthesized to form higher-level reflections that are dynamically retrieved to plan behavior. End-users can interact with a town of 25 generative agents using natural language. The architecture behind these generative agents is expected to be applied in collaborative scenarios.

\subsection{General purpose}
In addition to specific applications, some AI agents are developed for general purposes. These AI agents generally perform a wide range of tasks across diverse domains and attempt to reach the goal by thinking of tasks to do, executing them, and learning from the results. Auto-GPT\footnote{\url{https://github.com/Significant-Gravitas/Auto-GPT}} is one of the first examples of GPT-4 running fully autonomously. The feature of completing tasks autonomously without human intervention attracts people's attention. Similar to Auto-GPT, BabyAGI\footnote{\url{https://github.com/yoheinakajima/babyagi}} is a task-driven autonomous AI agent. BabyAGI constructs a task list dedicated to achieving the goal, derives further tasks based on the previous results, and executes these tasks in order of priority until the overall goal is achieved. Moreover, SuperAGI\footnote{\url{https://github.com/TransformerOptimus/SuperAGI}} and AgentGPT\footnote{\url{https://github.com/reworkd/AgentGPT}} support the building and deployment of autonomous AI agents, and have it embark on any goal imaginable. Although these AI agents are not so perfect and even have some deficiencies, their presentation is certainly an important step towards artificial general intelligence.

\subsection{Vision-Language model-based agent application}
LLM has already demonstrated outstanding capabilities in language-only scenarios. However, in some application scenarios, agents need to deal with multi-modal information, especially vision-language modalities. In such cases, modeling only the language information may not achieve satisfactory performance. Recent work considers equipping agents with the Vision-language model (VLM) to handle multi-modal information. In this subsection, we introduce some latest VLM-based agent applications. 
Some works attempt to apply VLM in the field of embodied AI and robotics that are based on visual and language modalities. For example, Khandelwal et al.~\cite{khandelwal2022simple} introduced CLIP~\cite{radford2021learning} into Embodied Agents, and demonstrated that CLIP can effectively enhance the task performance of embodied AI.
Driess et al.~\cite{Driess2023palme} combined ViT and PaLM to construct a multi-modal model named PaLM-E, which is applied in embodied reasoning. PaLM-E takes a multi-modal sequence (i.e., text and image) as input and converts it into text and image embeddings. Specifically, the image embedding is generated by the ViT and a projector encode images. Then, the text and image embeddings serve as input to PaLM for inferring the decisions that the robot needs to execute. Finally, the decisions are transformed into actions by a low-level policy or planner.
Some works focus on the navigation task. For instance, Dorbala et al.~\cite{dorbala2022clip} first used GPT-3 to break down navigation instructions into a series of sub-instructions. Then, at each time step, they utilized CLIP to select an image from the current panoramic view that corresponded to the sub-instructions, serving as the direction for the next navigation step. This process continued until the agent reached its target location. ZSON~\cite{majumdar2022zson} is an object-goal navigation agent designed to locate specific objects within an environment.
Besides, some works consider applied LVM in the field of multi-model conversational. For example, Video-ChatGPT~\cite{maaz2023video} is a video-based conversational agent fine-tuned using video instruction data. It first employs the visual encoder from CLIP to encode video frames into temporal and spatial features. Then, it utilizes a trainable adapter to map these features into the language space and combines them with query representations as inputs of LLM to generate responses. Li et al.\cite{li2023llava} introduce a conversational assistant for the biomedical field, named LLaVA-Med. It is continuously trained by LLaVA on multimodal biomedical datasets.

\section{Benchmarking}
Recently, LLM-based AI agents have attracted significant research interest. In order to evaluate the performance of the proposed agents, some works focus on designing more suitable benchmarks. For example, 
Valmeekam et al.~\cite{valmeekam2023planning} focused on assessing the planning ability of LLMs, which is a key component of AI agents.
Liu et al.~\cite{liu2023bolaa} designed a benchmark based on the WebShop and HotPotQA environment. Their goal is to compare the performance of multiple agent architectures equipped with different LLMs. 
Li et al.~\cite{li2023api} constructed a benchmark, named API Bank, to evaluate the ability of LLMs to use tools.   
Fan et al.~\cite{fan2022minedojo} proposed a simulator based on Minecraft to assess the performance of open-ended embodied agent.
Xu et al.~\cite{xu2023gentopia} designed a benchmark, named GentBench, which consists of public and private sections, with the aim of comprehensively evaluating the performance of agents. Specifically, GentBench includes a series of complex tasks that promote LLMs to employ external tools for addressing these challenges.
Banerjee~\cite{banerjee2023benchmarking} introduced an end-to-end benchmark that evaluates the performance of LLM-based chatbots by comparing generated answers with the gold answer.
Lin et al.~\cite{lin2023agentsims} presented a task-based evaluation method, which assesses the capabilities of agents based on their task completion within the interactive environment.
Liu et al.~\cite{liu2023agentbench} introduced a multi-dimensional benchmark, named AgentBench, which evaluates the performance of LLM across multiple environments.

\section{Conclusion}
In this paper, we presented a comprehensive and systematic survey of the LLM-based agents. We first introduced the difference between agents based on LLM and traditional methods, then reviewed the related works from the perspectives of components and application of AI agents. Furthermore, we have explored some pressing issues that require solutions and valuable research directions. With the development of LLM, an increasing amount of research attention has been directed toward the field of AI agents, resulting in the emergence of numerous new technologies and methods. Through this review, we aim to assist readers in swiftly grasping the key information and applications of AI agents, and also provide insights into future research directions.

\section{Bibliographical References}\label{sec:reference}

\bibliographystyle{lrec-coling2024-natbib}
\bibliography{reference.bib}

\end{document}